\documentclass[letterpaper, 10 pt, conference]{ieeeconf}  % Comment this line out if you need a4paper

\IEEEoverridecommandlockouts                              % This command is only needed if 
\title{\LARGE \bf
\methodname{}: An Inference Pipeline of Mass Estimation \\
for Vision-Language Models
}

\author{
Hisayuki Yokomizo\textsuperscript{1} \quad
Taiki Miyanishi\textsuperscript{1} \quad
Yan Gang\textsuperscript{1} \quad
Shuhei Kurita\textsuperscript{2,3} \quad
Nakamasa Inoue\textsuperscript{2} \quad
Yusuke Iwasawa\textsuperscript{1}
\\[2ex]
\textsuperscript{1}The University of Tokyo \quad
\textsuperscript{2}Institute of Science Tokyo  \quad
\textsuperscript{3}National Institute of Informatics \quad
\\[2ex]\\
}

\usepackage[style=numeric,citestyle=numeric,maxbibnames=999,maxcitenames=1,natbib=true,backend=biber, sortcites=true]{biblatex}
\bibliography{references.bib}
\usepackage{booktabs} 
\usepackage{xcolor}
\usepackage{amssymb}
\usepackage{hyperref}
\usepackage{graphicx}
\usepackage{cuted}
\usepackage{capt-of}
\usepackage{graphicx}
\usepackage{lipsum}
\usepackage{pifont}
\usepackage{pgffor}
\usepackage{amsmath}
\usepackage{xspace}
\usepackage{multirow}
\usepackage[table]{xcolor} 

\usepackage[most]{tcolorbox}

\definecolor{mygreen}{RGB}{255, 171, 64}

\newtcolorbox{promptbox}[1]{
    enhanced,
    colback=white,            % ボックス内の背景色
    colframe=mygreen,         % 枠線の色
    colbacktitle=mygreen,     % タイトル部分の背景色
    coltitle=white,           % タイトルの文字色
    fonttitle=\bfseries\normalsize,    
    title={#1},               % タイトルのテキスト（引数として受け取る）
    boxrule=1.5pt,            % 枠線の太さ
    arc=2mm,                  % 角の丸み
    auto outer arc,
    left=3mm, right=3mm, top=2mm, bottom=2mm, % 内側の余白
}

\definecolor{headergray}{gray}{0.9}
\definecolor{sectiongray}{gray}{0.95}
\definecolor{ourrow}{RGB}{230,245,255}

\newcommand{\methodname}{PhysQuantAgent\xspace}
\newcommand{\dataset}{VisPhysQuant}

\begin{document}

\maketitle
\thispagestyle{empty}
\pagestyle{empty}

%%%%%%%%%%%%%%%%%%%%%%%%%%%%%%%%%%%%%%%%%%%%%%%%%%%%%%%%%%%%%%%%%%%%%%%%%%%%%%%%

\begin{abstract}
Vision-Language Models (VLMs) are increasingly applied to robotic perception and manipulation,
yet their ability to infer physical properties required for manipulation remains limited. 
In particular, estimating the mass of real-world objects is essential for determining appropriate grasp force and ensuring safe interaction. 
However, current VLMs lack reliable mass reasoning capabilities, and 
most existing benchmarks do not explicitly evaluate physical quantity estimation under realistic sensing conditions.
In this work, we propose \methodname{}, a framework for real-world object mass estimation using VLMs, together with \dataset{}, a new benchmark dataset for evaluation. 
\dataset{} consists of RGB-D videos of real objects captured from multiple viewpoints, annotated with precise mass measurements. 
To improve estimation accuracy, we introduce three visual prompting methods that enhance the input image with object detection, scale estimation, and cross-sectional image generation to help the model comprehend the size and internal structure of the target object. 
Experiments show that visual prompting significantly improves mass estimation accuracy on real-world data, suggesting the efficacy of integrating spatial reasoning with VLM knowledge for physical inference.

\end{abstract}

%%%%%%%%%%%%%%%%%%%%%%%%%%%%%%%%%%%%%%%%%%%%%%%%%%%%%%%%%%%%%%%%%%%%%%%%%%%%%%%%

\section{INTRODUCTION}
Vision--Language Models (VLMs) jointly process visual inputs and natural language instructions.
When integrated into robotic systems, they enable end-to-end perception-to-action pipelines~\cite{zitkovich2023rt, huang2023voxposer, 2023palm}.
However, safe physical interaction requires not only semantic understanding but also quantitative reasoning about object properties.

In robotic manipulation, selecting an appropriate gripping force is critical for safe and stable operation.
The required gripping force depends on the object’s mass, size, and material properties~\cite{2000robot}.
Although force sensors provide feedback after contact, the initial gripping force must be set before interaction.
If this initial force is too small, the object may slip or be dropped.
If it is too large, the object may deform or be damaged.
Accurate vision-based mass estimation before contact is therefore essential for reliable manipulation.
Recent approaches such as NeRF2Physics~\cite{zhai2024physical} and PUGS~\cite{shuai2025pugs} estimate object mass by coupling volumetric reconstruction from 3D geometry with material inference using Large-Language Models (LLMs) or VLMs.
These methods reconstruct object geometry to compute volume and then infer material properties to approximate density.
However, these methods rely on computationally intensive 3D reconstruction and multi-stage processing pipelines. 
Such requirements increase computational cost and system complexity, which limits the efficiency and scalability in practical robotic applications.
An alternative approach is to directly leverage the prior knowledge and spatial reasoning capability of VLMs to estimate mass from RGB-D observations.
Despite their strong reasoning capabilities, it remains unclear whether VLMs can reliably estimate object mass directly from visual observations.
Furthermore, most existing object datasets lack mass annotations~\cite{wu2023omniobject3d,reizenstein2021common,ahmadyan2021objectron}.
Even when mass labels are provided, they are often restricted to large, furniture-scale objects in existing datasets~\cite{collins2022abo}, rather than small objects encountered in robotic manipulation.
As a result, datasets for mass estimation of small, manipulation-relevant objects remain limited.

\begin{figure}
  \centering
    \includegraphics[width=\linewidth]{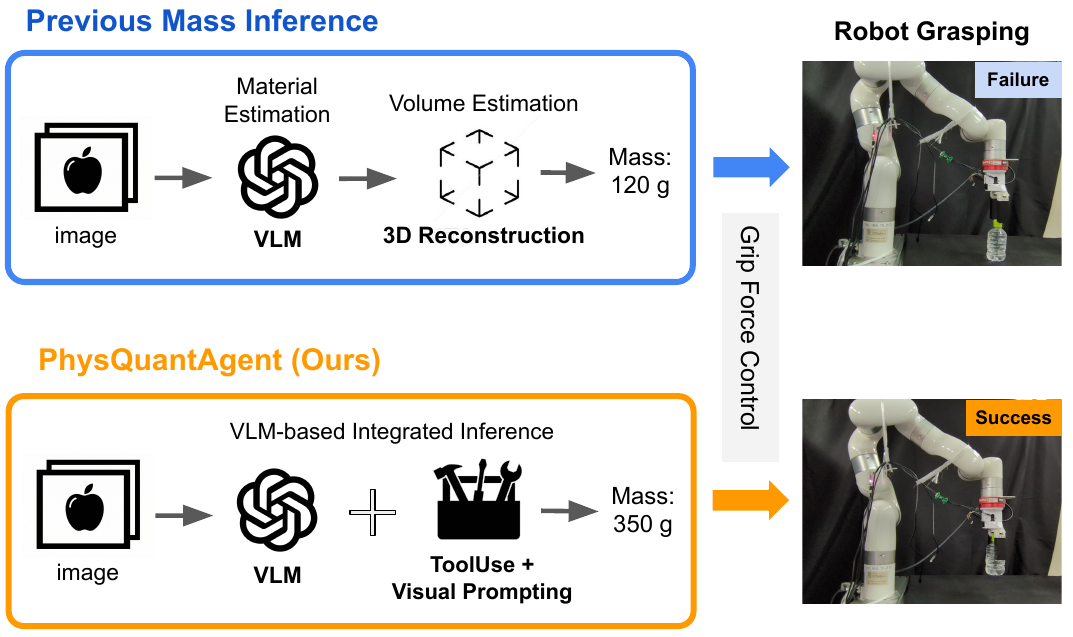}
  \caption{Comparison of previous mass estimation methods and our proposed {\methodname{}}. 
While prior approaches rely on computationally expensive 3D reconstruction from RGB-D images, {\methodname{}} directly infers object mass with a VLM, enabling fast estimation for grasp force control.}

  \label{fig:overview}
  \vspace{-0.4cm}
\end{figure}

To address these challenges, we propose \methodname{}, a plug-and-play inference pipeline that leverages visual prompting~\cite{wu2024visual} to enable efficient and reliable object mass estimation with VLMs. 
\autoref{fig:overview} compares previous mass estimation approaches with our proposed \textbf{\methodname{}}. 
Existing methods typically estimate object mass from RGB-D images through computationally expensive 3D reconstruction pipelines. 
In contrast, {\methodname{}} directly infers object mass using a VLM without requiring explicit 3D reconstruction, enabling fast and practical mass estimation. 
The estimated mass can then be used to adjust the grasp force during robotic manipulation. 
By explicitly providing scale and structural cues through visual prompts, \methodname{} helps VLMs better infer underlying 3D structure from 2D observations, which improves the estimation of physical quantities such as mass.

To evaluate the effectiveness of \methodname{} for visual mass estimation, 
we further introduce \textbf{\dataset{}}, a dataset of 360-degree RGB-D videos 
of small, robot-graspable everyday objects annotated with precise mass 
measurements. The dataset is designed to support systematic evaluation of 
vision-based mass estimation under realistic robotic manipulation settings. 
Unlike existing object datasets that primarily focus on large objects or 
lack physical annotations, \dataset{} provides ground-truth mass labels for 
objects commonly encountered in robotic grasping tasks.

Extensive experiments on \dataset{} systematically evaluate the mass inference capability of modern VLMs and show that visual prompting significantly improves estimation accuracy.
Furthermore, the results demonstrate that recent VLMs can outperform prior reconstruction-based approaches in visual mass estimation.
We further validate the practical effectiveness of \methodname{} through real-world robotic manipulation tasks, comparing it with prior mass-estimation methods.

\section{Related Work}

\subsection{Physical Properties Estimation}
Existing work on physical quantity estimation with VLMs has primarily focused on estimating object length.
For example, SpatialVLM~\cite{chen2024spatialvlm} constructs a pipeline that generates VQA-style training data from internet images to improve the quantitative reasoning ability of VLMs.
SpatialRGPT~\cite{cheng2024spatialrgpt} adds a module dedicated to depth images and trains them together with RGB images.
SpatialBot~\cite{cai2025spatialbot} incorporates depth information as textual inputs to VLMs, whereas SD-VLM~\cite{sd-vlm} encodes depth information directly during training.
Several benchmarks have been proposed to evaluate the spatial reasoning ability of VLMs~\cite{yu2025farvlmsvisualspatial, gholami2025spatialreasoning}. 
Beyond geometric quantities such as length, recent studies have also explored estimating physical properties such as reflectance, hardness, and surface roughness from tactile information~\cite{osada2024reflectance,yu2024octopi}.

As for mass estimation, studies have explored estimating object mass from images using deep neural networks~\cite{hamdan2019mass,10129573,nath2024mass}.
Recent approaches combine language models with 3D reconstruction techniques, such as NeRF or Gaussian Splatting, to estimate object volume and then infer material properties to approximate mass~\cite{zhai2024physical,shuai2025pugs}. 
However, these methods do not directly estimate the mass using language models and often suffer from large errors, with relative errors reported to be around 80--100\%.

\subsection{Physical Understanding Benchmarks}
Planning robotic manipulation with VLMs requires quantitative understanding of physical properties, such as object mass and material characteristics, in addition to reasoning about object interactions.
Existing benchmarks for physical understanding can be broadly categorized into two directions.
First, datasets such as CLEVRER~\cite{yi2019clevrer} and Super-CLEVER~\cite{li2023super} evaluate the ability to reason about object interactions, for example, predicting future motion after collisions.
Second, benchmarks including Physion++~\cite{tung2023physion++}, ContPhy~\cite{zheng2024contphy}, and PhysBench~\cite{chow2025physbench} incorporate tasks involving physical quantities.
However, these quantity-related tasks are primarily formulated as relative comparisons, 
rather than requiring estimation of absolute physical values.
While such relative reasoning evaluates qualitative physical understanding, it does not require estimating absolute physical values.
In contrast, robotic manipulation demands the estimation of absolute quantities, e.g., object mass, to determine appropriate grasp forces and ensure safe interaction.
This distinction highlights a fundamental gap between existing physical reasoning benchmarks and the quantitative requirements of real-world robotic manipulation.

\subsection{Object Datasets}
Various object datasets have been proposed as training or evaluation data for tasks such as 3D reconstruction or camera pose estimation. However, they present significant limitations when applied to the mass estimation task. 
\autoref{tab:dataset} summarizes representative object datasets and highlights their limitations for mass estimation, including missing mass annotations, limited viewpoint coverage, or reliance on synthetic 3D models.
For example, although Objaverse~\cite{deitke2023objaverse} is a large-scale object collection, its format is primarily restricted to 3D models. Objectron~\cite{ahmadyan2021objectron} provides real-world captures, yet its videos cover limited viewpoints, making it difficult to observe the full object structure.
Subsequent works have improved visual coverage through 360-degree captures~\cite{wu2023omniobject3d,reizenstein2021common} or depth integration~\cite{xia2024rgbd,guo2023handal}.
However, these datasets do not provide ground-truth mass annotations. Furthermore, even when scale information is introduced, such as in PhysXNet~\cite{cao2025physx}, it is often estimated by models rather than measured physical quantities.
Amazon Berkeley Objects (ABO)~\cite{collins2022abo} provides object size and mass annotations and has been used in prior mass estimation studies~\cite{zhai2024physical,shuai2025pugs}. 
However, the dataset primarily consists of product models and rendered images rather than real-world RGB-D captures, and its object categories are largely furniture-scale items. 
As a result, it contains relatively few small objects commonly encountered in robotic manipulation.
To address these limitations, we introduce \dataset{}, a dataset that pairs real-world RGB-D video captures with 360-degree object coverage and ground-truth mass annotations, enabling systematic evaluation of visual mass estimation in robotic manipulation scenarios.

% 赤いバツ印
%\newcommand{\rcross}{\textcolor{red}{\boldmath$\times$}}
\definecolor{myred}{RGB}{200,0,0}
\newcommand{\rcross}{\textcolor{myred}{\ding{55}}}
% 緑のボックスに入った白いチェックマーク
%\newcommand{\gcheck}{\textcolor{green}{\checkmark}}
\definecolor{mydarkgreen}{RGB}{0,140,0}
\newcommand{\gcheck}{\textcolor{mydarkgreen}{\checkmark}}

\begin{table}[t]
    \centering
    \caption{Comparison of existing datasets.}
    \label{tab:dataset}
    \resizebox{\linewidth}{!}{%
    \begin{tabular}{lcccccc}
    
        \toprule
        \textbf{Dataset} & \textbf{Video} & \textbf{View} & \textbf{Category} & \textbf{Real} & \textbf{Scale} & \textbf{Mass} \\ \hline

        % データ行
        OmniObject3D~\cite{wu2023omniobject3d} & RGB & Full & 190 & \gcheck & \rcross & \rcross \\ 
        CO3D~\cite{reizenstein2021common} & RGB & Full & 50 & \gcheck & \rcross & \rcross \\
        WildRGB-D~\cite{xia2024rgbd} & RGBD & Full & 44 & \gcheck & \rcross & \rcross \\ 
        Objectron~\cite{ahmadyan2021objectron} & RGB & Limited & 9 & \gcheck & \gcheck & \rcross \\
        HANDAL~\cite{guo2023handal} & RGBD & Limited & 17 & \gcheck & \gcheck & \rcross \\ 
        ABO~\cite{collins2022abo} & \rcross(3D) & -  & 63 & \rcross & \gcheck & \gcheck \\ 

        % 最後の行（太字）
%        \textbf{\dataset{} (Ours)} & RGBD & Full & xxx & \gcheck & \gcheck & \gcheck \\
        \textbf{\dataset{}} & RGBD & Full & 87 & \gcheck & \gcheck & \gcheck \\
        \bottomrule
    \end{tabular}
    }
\end{table}

\section{Method}
\subsection{Task Formulation}
We address vision-based object mass estimation for robotic manipulation with VLMs.
Given \(N\) multi-view RGB-D observations of a real-world object,
\(\mathcal{I}_0=\{(I_n, D_n)\}_{n=1}^{N}\),
where \(I_n\) and \(D_n\) denote the RGB image and corresponding depth map captured from viewpoint \(n\),
our goal is to estimate the object's physical mass \(m \in \mathbb{R}_{>0}\).
A VLM predicts the mass as
\begin{align}
\hat{m} = VLM(Q, \mathcal{I}_0),
\end{align}
where \(Q\) is a textual prompt and \(VLM(\cdot)\) denotes the VLM inference function.
We define the objective as minimizing the absolute mass error:
%\begin{align}
$\mathcal{L} = \lvert m - \hat{m} \rvert$.
%\end{align}
Since the task targets robot-arm manipulation, we assume the robot is equipped with a calibrated RGB-D sensor whose intrinsic parameters (e.g., focal length and sensor size) are known.
Together with metric depth measurements, this enables physically grounded scale reasoning for mass estimation.

\subsection{\methodname{}} 
\begin{figure}[t]
	\centering
	\includegraphics[width=\linewidth]{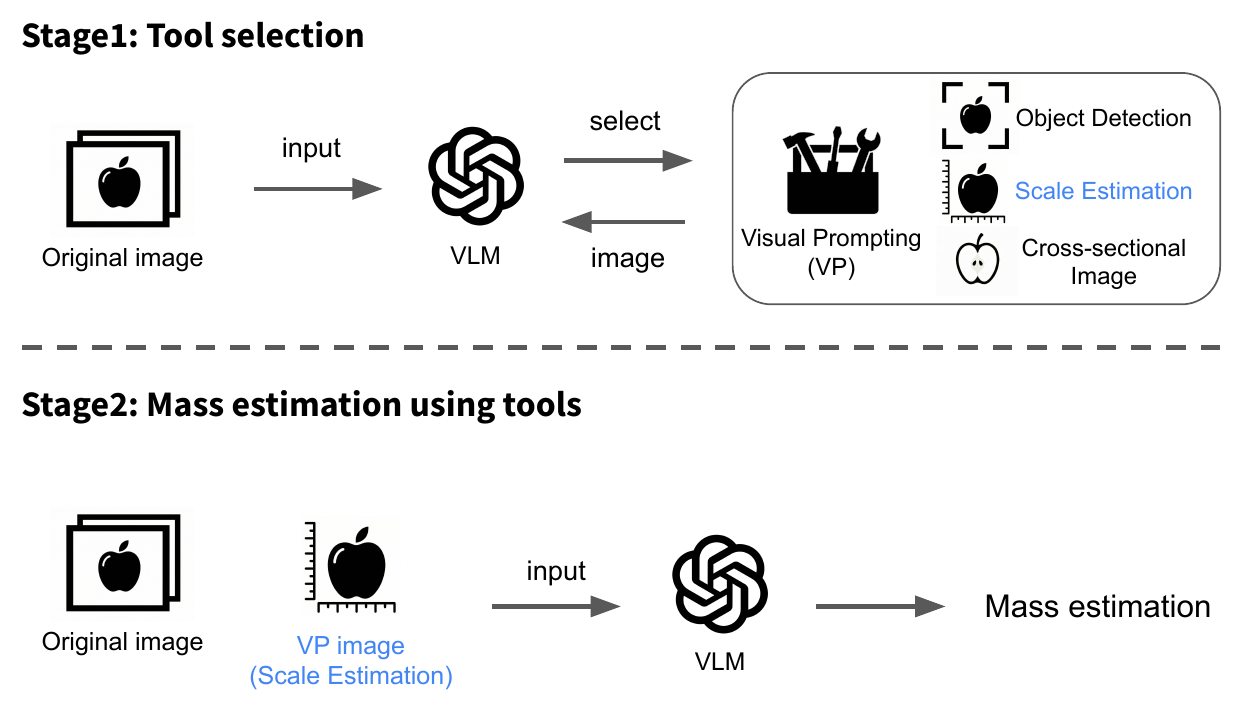}
	\caption{\textbf{\methodname{}} framework: First, the VLM selects the tool to use (Stage1), then estimates the mass of the object from the original image and the image with the tool applied (Stage2).}
	\label{fig:framework}
\end{figure}

%  提案手法の概要を書く
\noindent \textbf{Overview.} 
Mass estimation from raw RGB-D observations is fundamentally underconstrained. 
Although depth provides geometric information, it does not directly reveal physical cues such as volume, density, or calibrated scale. 
As a result, VLMs often rely on superficial visual cues, leading to unstable mass predictions.
To address this limitation, we propose \methodname{}, an inference framework that leverages Visual Prompting~\cite{wu2024visual} to make physically relevant cues explicit to the VLM (see \autoref{fig:framework}).
Rather than relying solely on unprocessed observations, \methodname{} generates structured visual augmentations that expose complementary attributes required for mass estimation.

\methodname{} comprises three Visual Prompting modules.
Each module is designed to emphasize a distinct physical aspect of the object that is informative for mass estimation.
Given an RGB-D observation, the VLM analyzes the scene and selects the most appropriate module for the target object.
The selected module produces an augmented image that provides relevant visual cues.
This adaptive cue selection mechanism improves the reliability of mass estimation.

In the inference pipeline, the VLM first selects the Visual Prompting, then returns the image with that applied, and the model infers mass using both images.
\autoref{fig:prompt_example} shows the example of input prompt in each stage.

\begin{figure}[htbp]
    \centering
    % --- ボックス (a) ---
    \begin{promptbox}{Stage1: Tool selection}
You will be given an image of an object. Your task is to estimate the weight of the object.\\
Choose which tools to use (you may choose multiple) to help estimate the object’s weight.\\
Available tools:\\
- Object Detection: create an annotated image with detected object bounding box.\\
- Scale Estimation: create an annotated image with measured X/Y lengths.\\
- Cross~sectional Image: generate three-view cross-sectional images of the object.
    \end{promptbox}
    
    \vspace{0.1em} % ボックス間の余白
    
    % --- ボックス (b) ---
    \begin{promptbox}{Stage2 Mass estimation}
You are a physics and engineering assistant. \\
Estimate the weight of the object from the images. \\
Reason step by step and finally state your answer in kilograms like 'Answer: - kg'.
    \end{promptbox}
    
    \caption{Example of the instruction prompt in \methodname{}.}
    \label{fig:prompt_example}
\end{figure}

\vspace{0.1cm}
\noindent \textbf{Mass Estimation Tools.} 
In order to improve the accuracy of mass inference by explicitly providing information such as the size and internal structure of an object to the VLM, we introduce three complementary Visual Prompting tools~\cite{wu2024visual}:
(a) object detection for target localization,
(b) scale estimation for geometric reasoning, and
(c) cross-sectional image generation for structural understanding.
The selected tool generates an augmented image that explicitly highlights a specific physical attribute of the object.

\vspace{0.1cm}
\noindent \textit{\textbf{(a) Object Detection}}:
Mass estimation must be conditioned on the correct target object.
Without explicit localization, a VLM may incorporate surrounding objects and produce biased estimates.
To prevent this, we localize the target using Grounding DINO~\cite{liu2024grounding} and provide the resulting bounding box to the VLM.
This visual constraint restricts attention to the object of interest and reduces background interference.

\vspace{0.1cm}
\noindent \textit{\textbf{(b) Scale Estimation}}:
Mass is directly related to physical size. However, pixel coordinates do not encode metric scale. We therefore overlay geometric axes with metric information to provide explicit scale cues.
Unlike 3DAxiesPrompts~\cite{liu20233daxiesprompts}, which requires manual annotation, our method automatically generates scale references.

\begin{figure}[t]
	\centering
    \includegraphics[width=\linewidth]{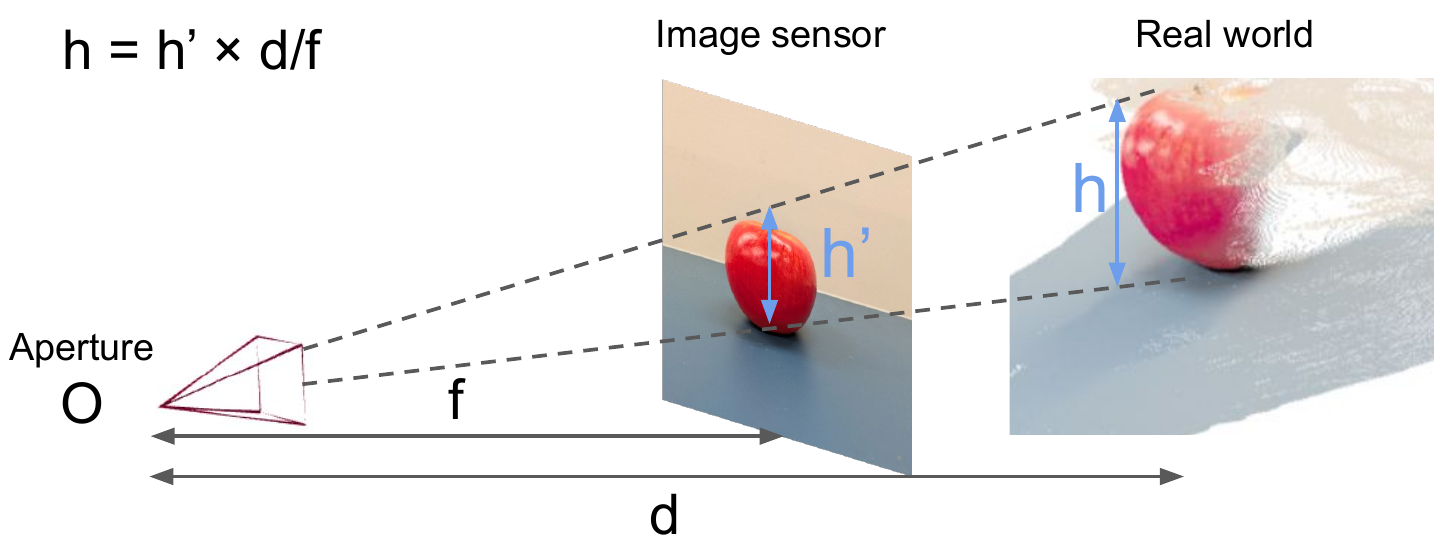}
	\caption{Pinhole camera model.}
	\label{fig:pinhole}
\end{figure}

\begin{figure}[t]
	\centering
	\includegraphics[width=\linewidth]{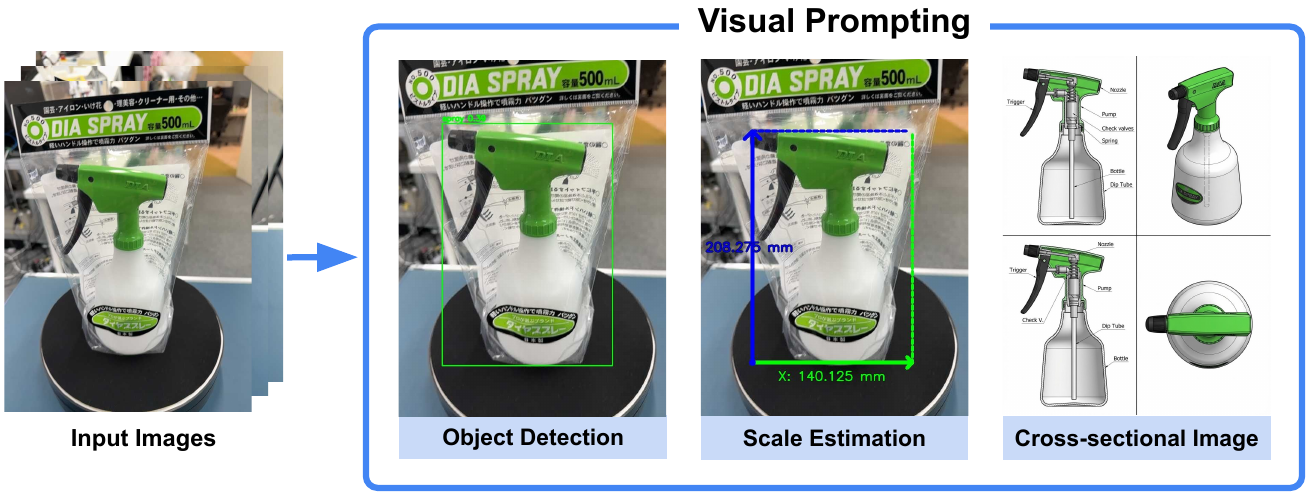}
	\caption{Examples of Visual Prompting.}
	\label{fig:example_vp}
\end{figure}

When RGB-D input is given, metric distances are computed from camera intrinsics and depth under the pinhole model (see \autoref{fig:pinhole}).
After segmenting the object, we estimate its horizontal and vertical extents and overlay the corresponding metric lengths as scale annotations (see \autoref{fig:example_vp}).

\vspace{0.1cm}
\noindent \textit{\textbf{(c) Cross-sectional Image}}:
External appearance alone does not determine mass, as objects with similar shapes may differ in internal material distribution and effective density.
To expose cues about internal occupancy, we generate cross-sectional visualizations using an image editing model (e.g., Nano Banana~\cite{nanobanana}).
Conditioned on the input image and text instructions, the model synthesizes sliced views from multiple directions, which are appended to the original observation to support density-aware mass reasoning.

\section{Dataset}
\begin{figure}[t]
  \centering
  \includegraphics[width=\linewidth]{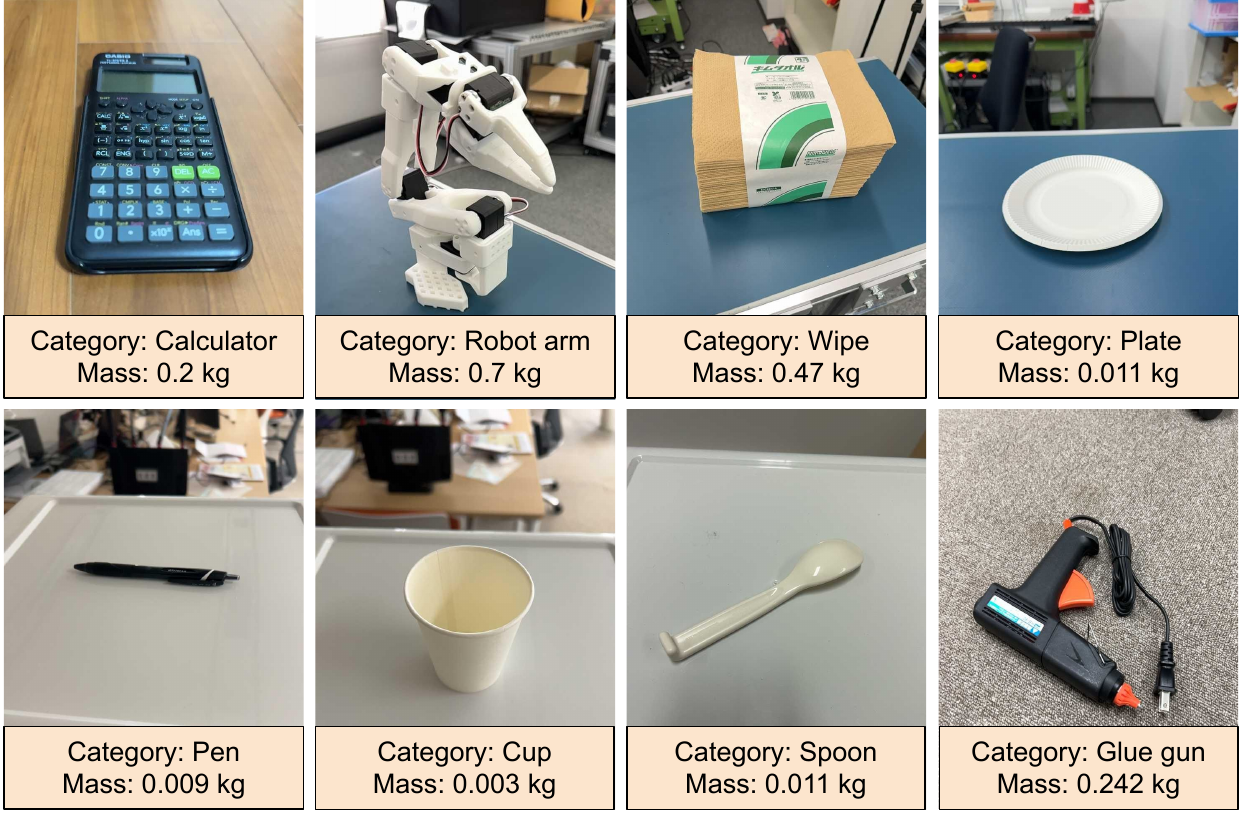}
  \caption{\textbf{\dataset{}} contains approximately 300 RGBD videos of objects with mass annotations. To ensure diversity, the data were collected in diverse background settings.}
  \label{fig:dataset}
\end{figure}

\dataset{} is a RGB-D object dataset for evaluating VLMs' mass estimation ability. 
Our dataset consists of approximately 300 samples, each annotated with ground-truth object mass as shown in \autoref{fig:dataset}.
Unlike other object datasets~\cite{collins2022abo, cao2025physx}, \dataset{} provides real-world RGB-D videos with ground-truth mass annotations. 
This makes it suitable for mass estimation in robotic manipulation scenarios.

\subsection{Data Collection}
We collected RGB-D videos using \textit{Record3D} on a LiDAR-equipped iPhone 16 pro.
The application provides synchronized RGB frames and metric depth maps reconstructed from onboard LiDAR measurements.
This setup enables direct acquisition of geometrically consistent RGB-D observations.

Each object was recorded by moving the camera around it to obtain near 360-degree coverage.
We maintained a moderate and consistent camera-to-object distance to ensure stable depth reconstruction.
Data were discarded if (i) substantial regions of the object were not reconstructed, (ii) severe camera shake degraded depth quality, or (iii) the object was partially outside the field of view.
These criteria ensure reliable geometric information for subsequent scale reasoning.

The ground-truth mass of each object was measured using a calibrated digital scale, \textit{TANITA KJ-212} (precision: $\pm 0.3$~g).
For objects that could not be weighed directly due to size or safety constraints, we used manufacturer-provided specifications.
In such cases, we verified consistency between catalog values and approximate manual measurements when possible.

Objects were placed individually or in moderately cluttered tabletop environments to reflect realistic household manipulation scenarios.
The backgrounds consisted of common indoor surfaces such as wood, plastic, and fabric under typical indoor lighting conditions. 
The camera-to-object distance ranged from approximately $0.3$ to $0.5$ meters to ensure stable RGB-D capture. 
Both isolated-object and cluttered-context recordings are included to evaluate robustness to contextual interference.

\subsection{Dataset Statistics}

\begin{figure}[t]
	\centering
	\includegraphics[width=\linewidth]{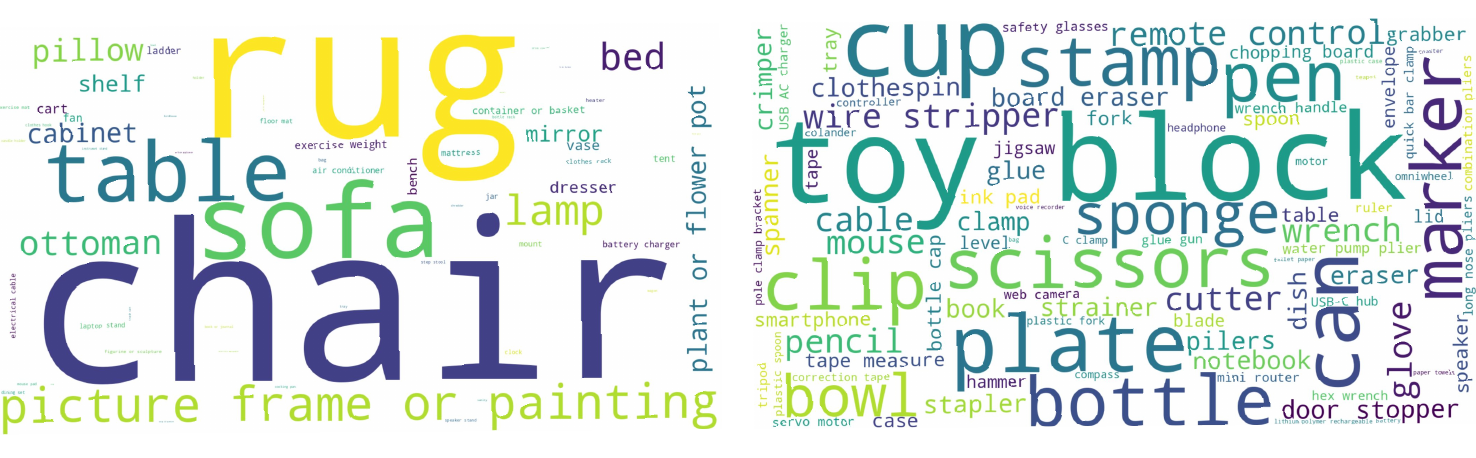}
	\caption{A comparison of category distribution of ABO~\cite{collins2022abo} (left) and \dataset{} (right)}
	\label{fig:category}
\end{figure}

We analyze the category distribution of \dataset{} and compare it with ABO~\cite{collins2022abo}, a widely used object dataset that has been adopted in prior mass estimation studies~\cite{zhai2024physical,shuai2025pugs}. 
\autoref{fig:category} shows the category distributions of the two datasets. 
ABO is furniture-centric and exhibits limited object diversity, whereas \dataset{} contains a broader range of objects that can be grasped by humans or robotic manipulators.

\begin{figure}[t]
	\centering
	\includegraphics[width=\linewidth]{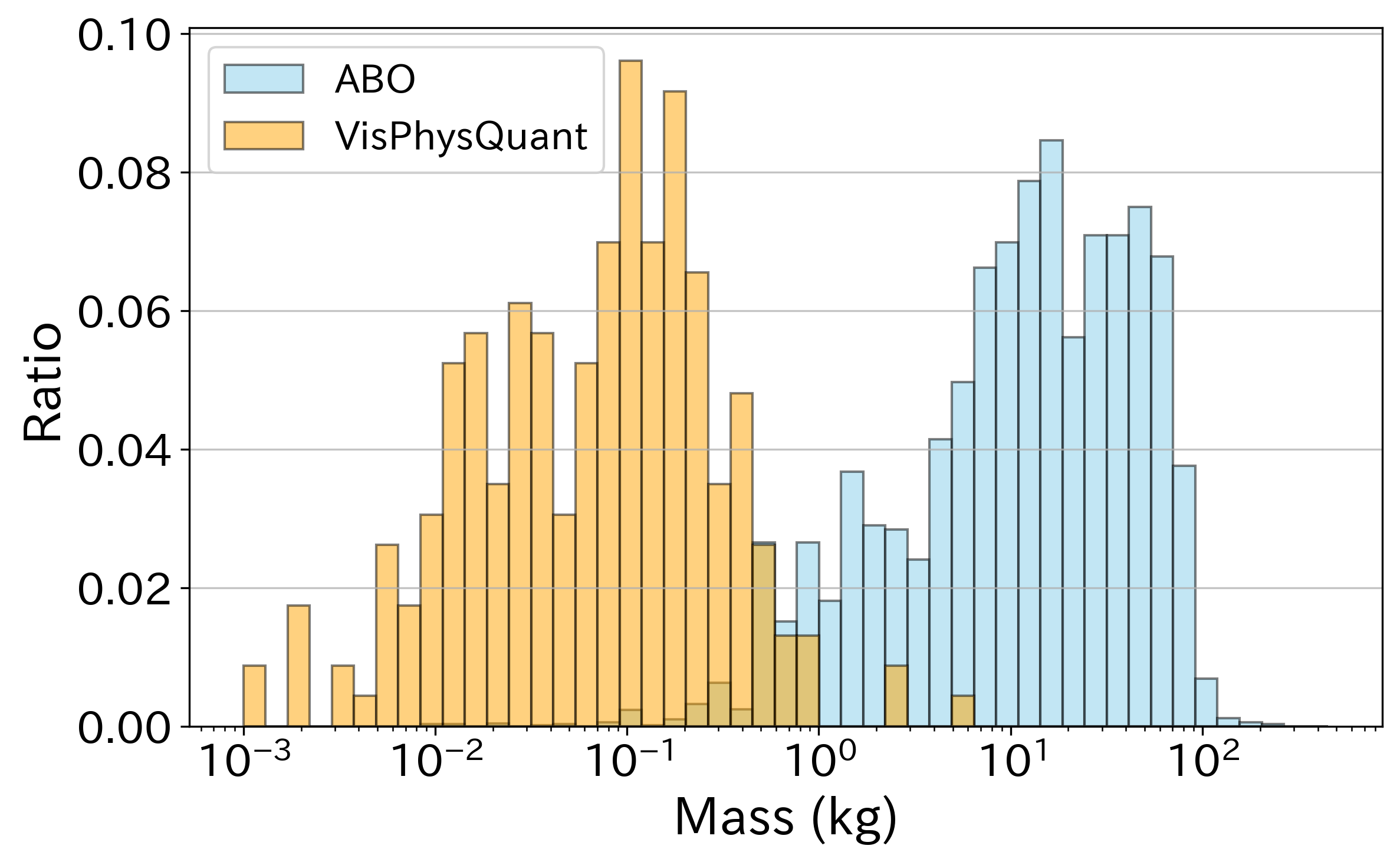}
	\caption{A comparison of weight distribution of ABO~\cite{collins2022abo} and \dataset{}.}
	\label{fig:mass_distribution}
\end{figure}

\autoref{fig:mass_distribution} compares the mass distributions of the two datasets. 
ABO primarily contains furniture-scale objects with a median mass of $12.7$~kg. 
In contrast, \dataset{} focuses on small- and medium-sized objects that are suitable for robotic manipulation and enables systematic evaluation of mass estimation within realistic manipulation ranges. 
The mass range of \dataset{} spans from approximately $0.001$~kg to $5$~kg, with a median of $0.08$~kg. 

This range aligns with the payload capacity of common household and research robotic manipulators, reflecting realistic manipulation tasks such as grasping and pick-and-place.
Notably, \dataset{} also includes objects below $100$~g, which are underrepresented in existing large-scale 3D object datasets but frequently appear in everyday manipulation scenarios.

\section{Experiment}
%\subsection{Benchmarks and metrics.} 
%\subsection{Setup}
\subsection{Experimental Setup}
\noindent \textbf{Benchmarks and Metrics.} 
We evaluate VLMs on the mass estimation task using \dataset{}.
We use Minimum Ratio Error (MnRE) as a metric, following prior work~\cite{shuai2025pugs,zhai2024physical}.
MnRE is defined as
\begin{align}
\mathrm{MnRE} = \min\left(\frac{m}{\hat{m}}, \frac{\hat{m}}{m}\right),
\end{align}
where $m$ denotes the ground-truth mass and $\hat{m}$ the predicted mass.
MnRE provides a symmetric and scale-invariant measure that penalizes underestimation and overestimation equally.
Compared with absolute or relative error, MnRE avoids bias toward overestimation or underestimation and provides a more balanced evaluation of mass prediction accuracy.

\vspace{0.1cm}
\noindent \textbf{Baselines.} 
We compare \methodname{} with NeRF2Physics~\cite{zhai2024physical}, a reconstruction-based approach for mass estimation. 
We also evaluate VLM baselines without visual prompting, including Qwen3-VL-8B, Gemini~2.5~pro and Gemini~3.1~pro~\cite{bai2025qwen3,gemini2.5,gemini3.1}. 
Although we also considered PUGS~\cite{shuai2025pugs}, preliminary experiments with the publicly available implementation on the ABO dataset showed lower performance than NeRF2Physics, and we therefore adopt NeRF2Physics as the reconstruction-based baseline.

\vspace{0.1cm}
\noindent \textbf{Implementation Details.}
To process videos with VLMs, we extract sequential frames. Specifically, we sample frames every 30 frames from the 15-second, 30-fps videos, resulting in approximately 15 images per video.
The image generation model used was Gemini 3 Pro Image-preview (Nano Banana Pro~\cite{nanobanana}). For object detection and length estimation, we use Grounded-Segment-Anything~\cite{ren2024grounded}.

\subsection{Experimental Results}

\begin{figure}
	\centering
    \includegraphics[width=\linewidth]{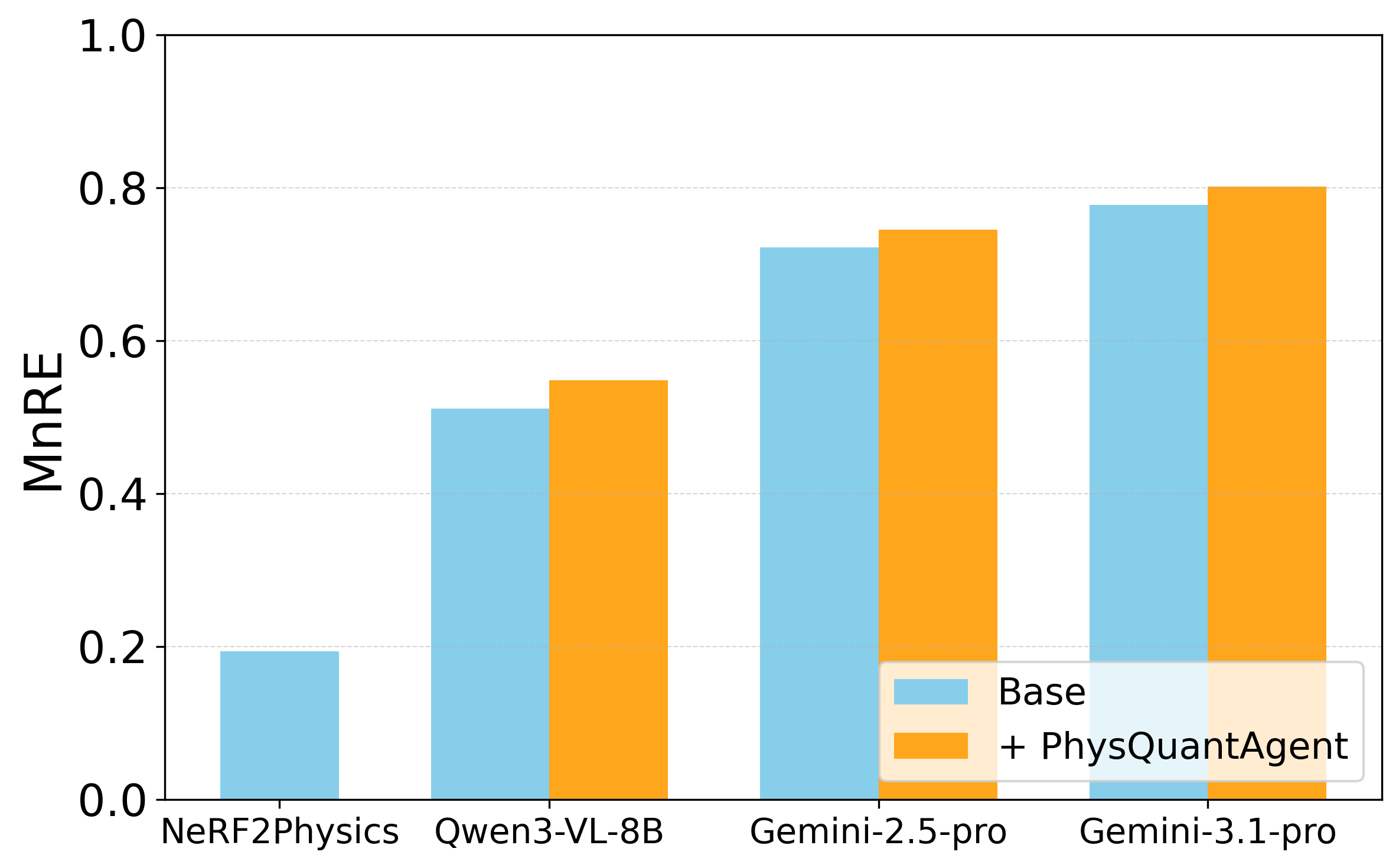}
	\caption{The mass estimation performance of NeRF2Physics and VLMs with \methodname{}.}
	\label{fig:basic_result}
\end{figure}

\vspace{0.1cm}
\noindent \textbf{Overall Performance.}
\autoref{fig:basic_result} presents the mass estimation results of VLMs and NeRF2Physics on \dataset{} using MnRE as the evaluation metric. 
Recent VLMs outperform the reconstruction-based method NeRF2Physics even without explicit volume estimation.
Applying \methodname{} consistently improves the MnRE of each tested VLM, demonstrating the effectiveness of the proposed visual prompting.

\vspace{0.1cm}
\noindent \textbf{Ablation Study.}
\autoref{tab:vp_ablation} presents the ablation study evaluating the contribution of each visual prompting module.
All prompting strategies improve the MnRE compared with the baseline without visual prompting, indicating that providing explicit spatial or structural cues helps VLMs produce more reliable mass estimates.

\begin{figure}[t]
	\centering
    \includegraphics[width=\linewidth]{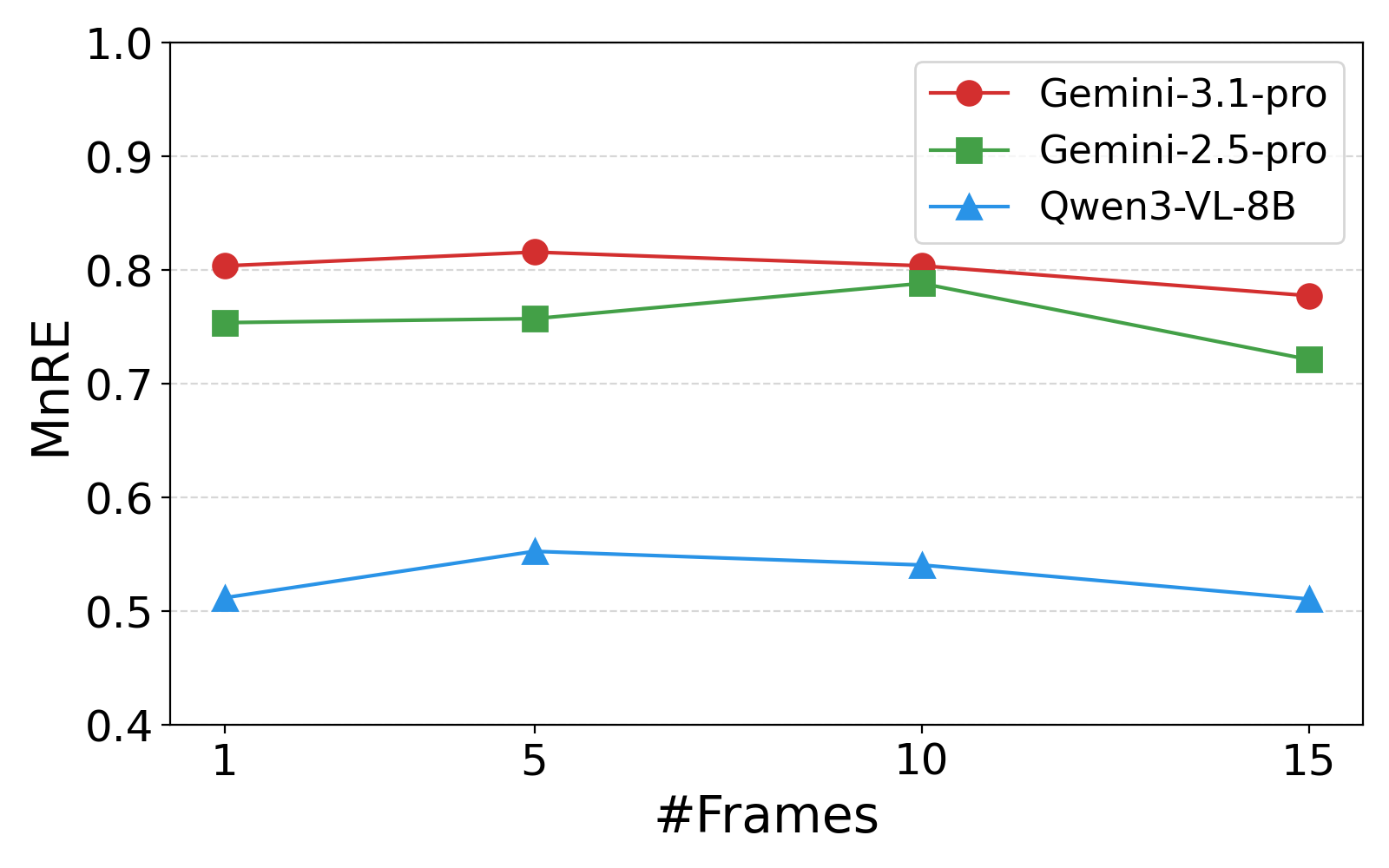}
	\caption{Ablation on input frame count.}
	\label{fig:frame_ablation}
\end{figure}

\autoref{fig:frame_ablation} shows the effect of varying the number of input frames.
The result shows that accuracy does not increase monotonically with more frames.
Each of the tested VLMs achieve the best performance with approximately 5--10 frames.
Using too few frames provides insufficient structural information, whereas too many frames introduce redundant observations that slightly degrade inference accuracy.
Notably, \methodname{} achieves competitive performance with only a few frames, while reconstruction-based methods such as NeRF2Physics typically require dozens of images (around 30) for volume estimation.

\begin{figure}[t]
	\centering
    \includegraphics[width=\linewidth]{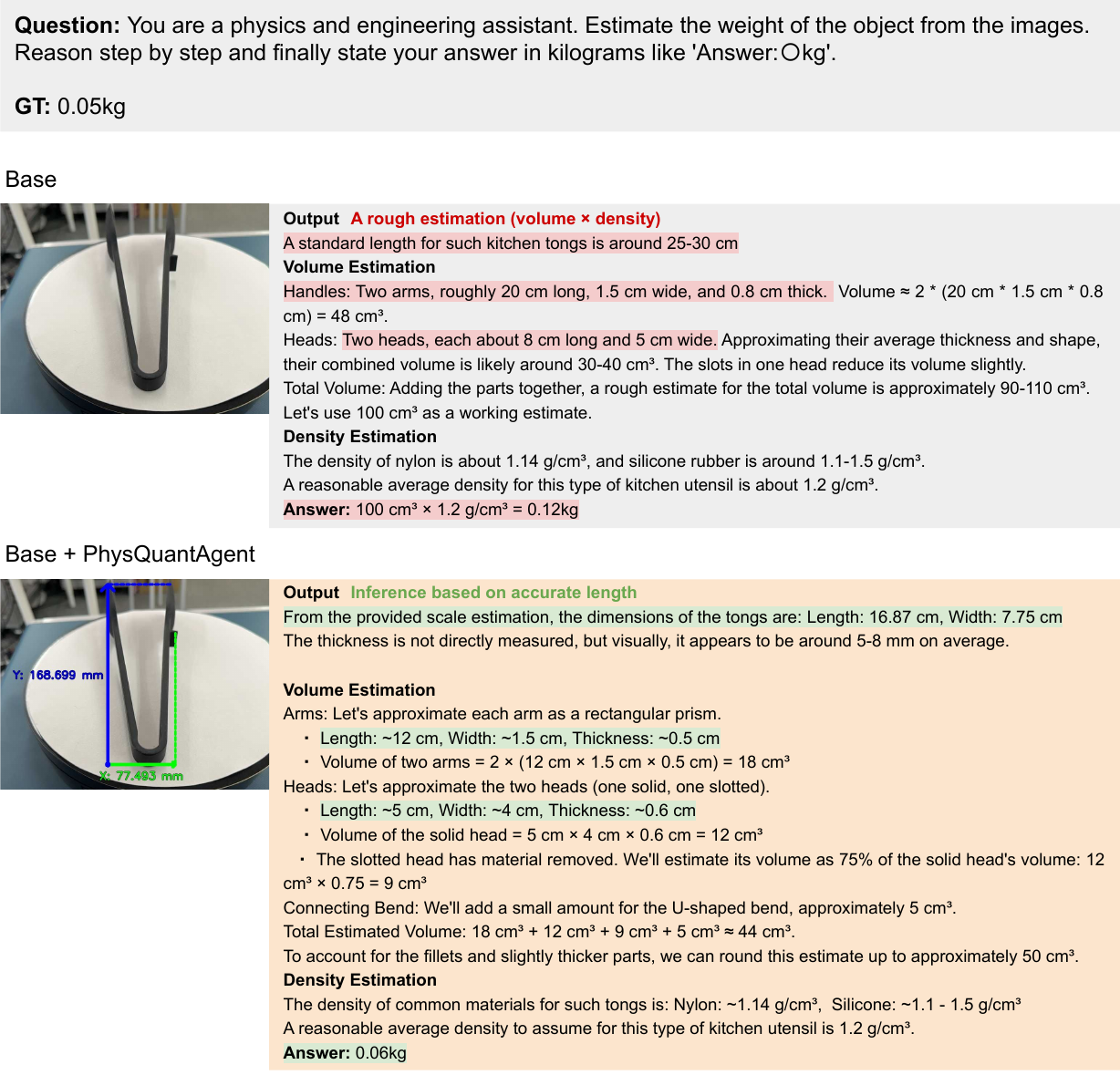}
	\caption{Qualitative example of \methodname{}.}
	\label{fig:good_example}
\end{figure}

\begin{figure}[t]
	\centering
    \includegraphics[width=\linewidth]{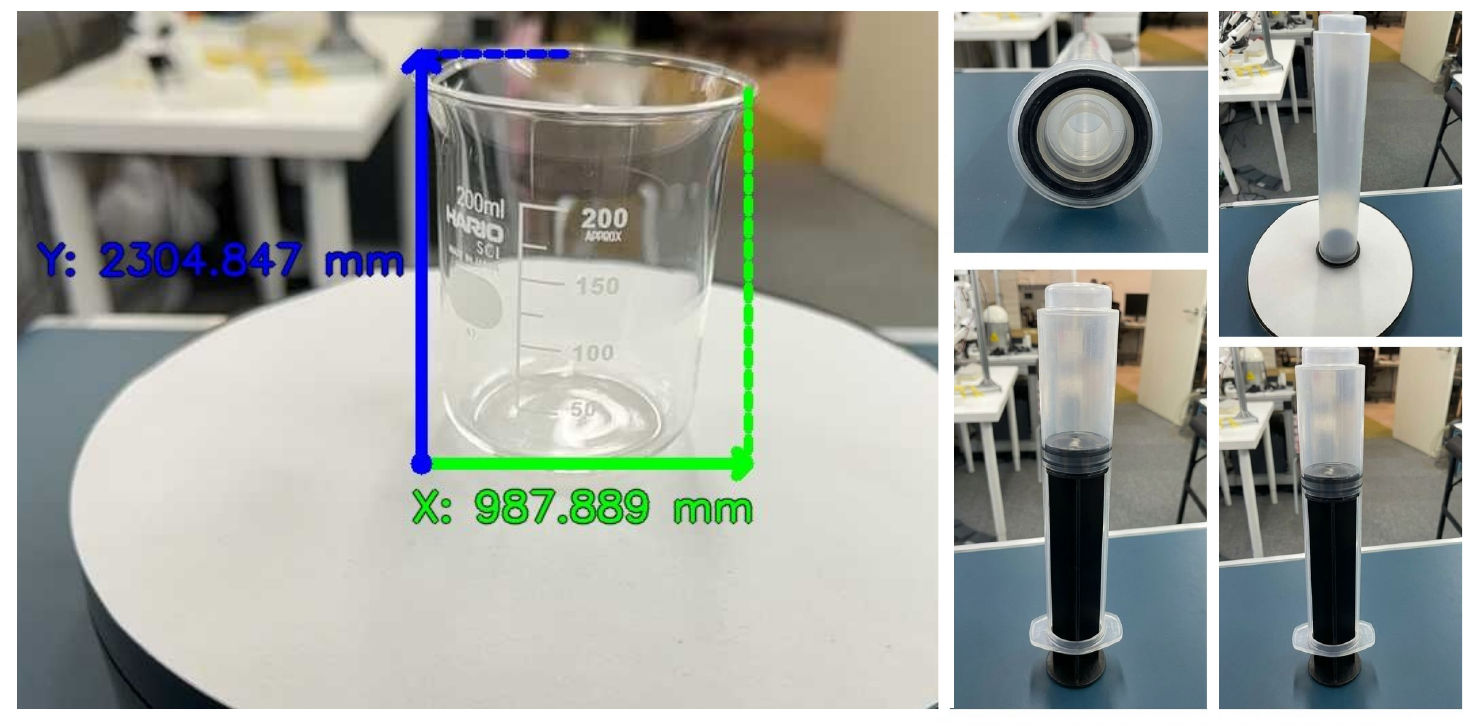}
	\caption{Error cases in Visual Prompting. Scale estimation can yield large errors for transparent objects (left), while Cross-sectional Image generation is prone to hallucinations, such as the synthesis of non-existent objects (right).}
	\label{fig:bad_example}
\end{figure}

\noindent \textbf{Qualitative Results.}
\autoref{fig:good_example} shows an example where visual prompting significantly improves mass estimation accuracy. 
By using scale estimation, the VLM can infer an object's size and volume, which leads to more accurate mass predictions.
On the other hand, visual prompting can also introduce errors, as shown in \autoref{fig:bad_example}. 
Distance is estimated using depth measurements obtained from a LiDAR sensor based on the time-of-flight of emitted laser pulses. 
However, LiDAR signals can penetrate transparent objects such as glass, resulting in overestimated depth values. 
In such cases, depth estimation models such as Depth Anything~\cite{yang2024depth} could help mitigate this issue.
Cross-sectional image generation sometimes also fail to produce appropriate structural references. 
The generated images sometimes contain artifacts or non-existent objects, which can lead to overestimation of object mass.

\begin{table}[t]
\centering
\setlength{\tabcolsep}{6pt}
\caption{Ablation of Visual Prompts.}
\resizebox{\columnwidth}{!}{
\begin{tabular}{llc}
\toprule
\textbf{Base Model} & \textbf{Visual Prompt} & \textbf{MnRE} $\uparrow$ \\
\midrule
\multirow{4}{*}{Gemini-2.5-pro}
 & Baseline              & 0.721 \\
 & \; + Object Detection      & 0.747 \\
 & \; + Scale Estimation      & 0.754 \\
 & \; + Cross-sectional Image & 0.738 \\
\midrule
\multirow{4}{*}{Gemini-3.1-pro}
 & Baseline               & 0.778 \\
 & \; + Object Detection      & 0.806 \\
 & \; + Scale Estimation      & 0.789 \\
 & \; + Cross-sectional Image & 0.796 \\
\bottomrule
\end{tabular}
}
%\caption{Effect of different visual prompting strategies on MnRE. Higher is better ($\uparrow$).}
\label{tab:vp_ablation}
\end{table}

\subsection{Robot Application}

\begin{figure}[t]
	\centering
    \includegraphics[width=0.7\linewidth]{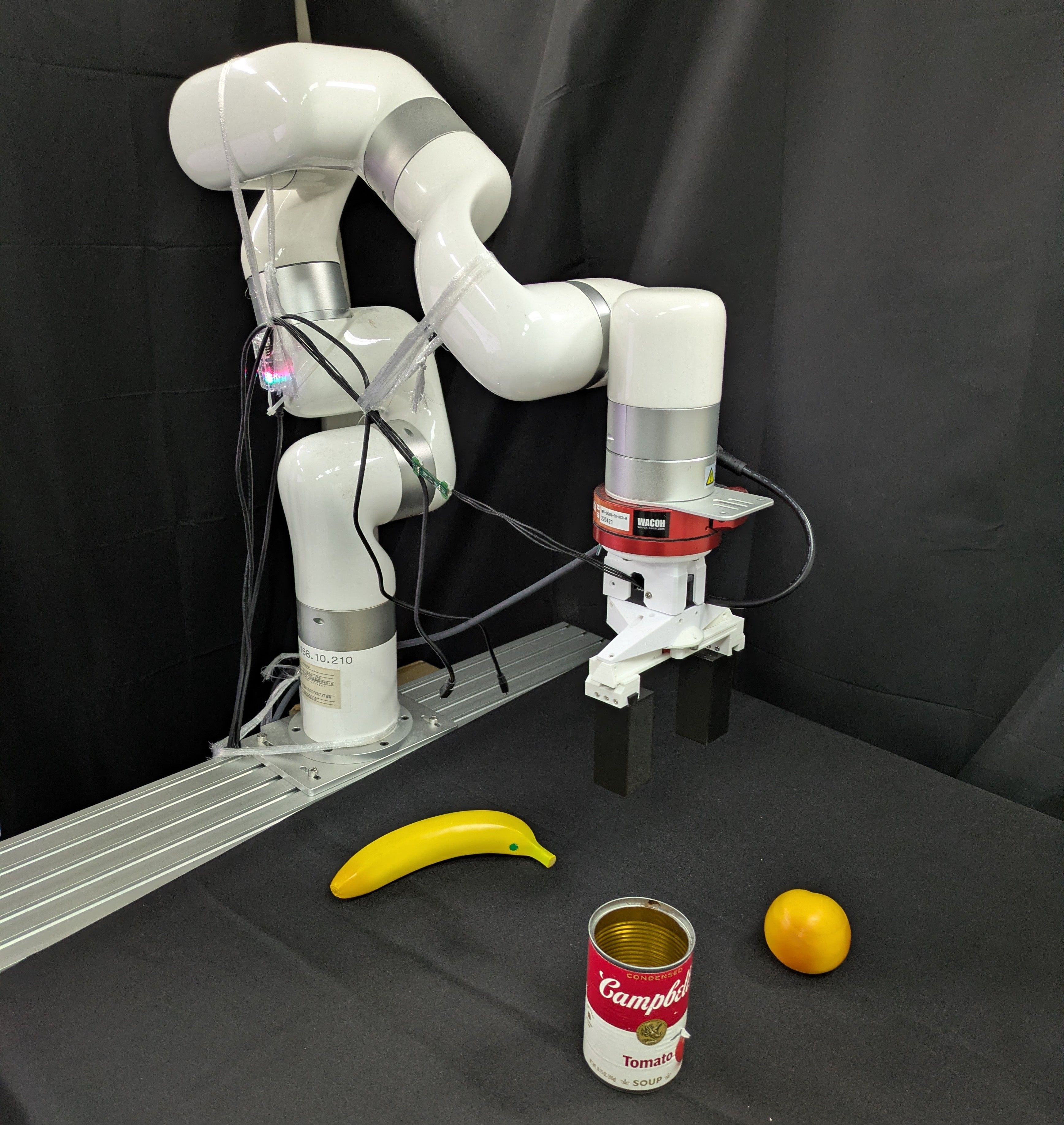}
	\caption{Example of a robotic manipulation task.}
	\label{fig:robot}
\end{figure}

We evaluate the practical utility of \methodname{} in a robotic manipulation task using the \textit{xArm7}. 
\autoref{fig:robot} demonstrates this robotic manipulation scenario.
First, we record a video of the target object and estimate its mass using mass estimation methods.
Based on the estimated mass, the gripping force of the robot arm is adjusted by modulating the applied current. 
The gripper width is automatically determined by gradually closing the gripper until contact with the object is detected.
As shown in \autoref{fig:overview}, inaccurate mass estimation from NeRF2Physics leads to insufficient gripping force, causing the robot to fail to lift the object. 
In contrast, \methodname{} provides more reliable mass estimates, enabling stable robotic grasping.

\section{Conclusion}
We propose \methodname{}, a plug-and-play inference pipeline for object mass estimation that integrates tool-based visual prompting with VLM reasoning.
Our results demonstrate that explicitly providing spatial cues enables VLMs to perform more reliable physical inference. 
We also introduced \dataset{}, a dataset of real-world RGB-D videos of small, robot-graspable everyday objects annotated with ground-truth mass measurements, which provides a benchmark for evaluating visual mass estimation.
The proposed method and dataset enable more reliable estimation of object mass from visual observations, facilitating grasp-force adjustment in robotic manipulation.


%%%%%%%%%%%%%%%%%%%%%%%%%%%%%%%%%%%%%%%%%%%%%%%%%%%%%%%%%%%%%%%%%%%%%%%%%%%%%%%%

%%%%%%%%%%%%%%%%%%%%%%%%%%%%%%%%%%%%%%%%%%%%%%%%%%%%%%%%%%%%%%%%%%%%%%%%%%%%%%%%

%%%%%%%%%%%%%%%%%%%%%%%%%%%%%%%%%%%%%%%%%%%%%%%%%%%%%%%%%%%%%%%%%%%%%%%%%%%%%%%%
%\section*{APPENDIX}


%\section*{ACKNOWLEDGMENT}

%he preferred spelling of the word ÒacknowledgmentÓ in America is without an ÒeÓ after the ÒgÓ. Avoid the stilted expression, ÒOne of us (R. B. G.) thanks . . .Ó  Instead, try ÒR. B. G. thanksÓ. Put sponsor acknowledgments in the unnumbered footnote on the first page.

%%%%%%%%%%%%%%%%%%%%%%%%%%%%%%%%%%%%%%%%%%%%%%%%%%%%%%%%%%%%%%%%%%%%%%%%%%%%%%%%

%References are important to the reader; therefore, each citation must be complete and correct. If at all possible, references should be commonly available publications.

\renewcommand*{\bibfont}{\footnotesize}

{\footnotesize
\printbibliography
}

\end{document}